# Active stereo vision three-dimensional reconstruction by RGB dot pattern projection and ray intersection

YongCan Shuang and ZhenZhou Wang *

College of Electrical and Electronic Engineering, Shandong University of Technology, Zibo 255000, China; 14110402154@stumail.sdut.edu.cn (Y.S.); zzwangsia@yahoo.com (Z.W.);

* Correspondence: zzwangsia@yahoo.com;

**Abstract:** Active stereo vision is important in reconstructing objects without obvious textures. However, it is still very challenging to extract and match the projected patterns from two camera views automatically and robustly. In this paper, we propose a new pattern extraction method and a new stereo vision matching method based on our novel structured light pattern. Instead of using the widely used 2D disparity to calculate the depths of the objects, we use the ray intersection to compute the 3D shapes directly. Experimental results showed that the proposed approach could reconstruct the 3D shape of the object significantly more robustly than state of the art methods that include the widely used disparity based active stereo vision method, the time of flight method and the structured light method. In addition, experimental results also showed that the proposed approach could reconstruct the 3D motions of the dynamic shapes robustly.

## 1. Introduction

Stereo vision technique calculates the three-dimensional shape by finding the corresponding points between the left camera and the right camera. The most challenging part of stereo vision is to match two views of the cameras robustly. Up to date, many stereo matching methods based on working different principles have been proposed. However, none of them really solved the problem at large. To overcome the difficulties of matching texture-less objects, structured light illumination [1-11] has been used to determine the correspondences between the two views. Up to date, so many structured light patterns [1] have been designed to overcome the correspondence problems in three-dimensional imaging and all the structured light patterns could be used directly for active stereo vision. Each designed structured light pattern has its advantages and disadvantages. For instance, the laser speckle pattern is used to establish the stereo vision correspondences in [4-5]. The advantages of the laser speckle pattern based active stereo vision include the relatively high reconstruction resolution and the high resistance to external light interference. The disadvantage of the laser speckle pattern might be its relatively low accuracy. As reported in [6], the laser speckle based active stereo vision is less accurate than the stripe pattern based active stereo vision. Another popular structured light pattern is the phase pattern that has been used in [7] to establish the stereo vision correspondences between two camera views. The advantages of the phase pattern based active stereo vision are its high resolution and its high accuracy. The disadvantages of the phase pattern based active stereo vision are its low resistance to external light interference, its difficulty to establish exact correspondences between two cameras views and its low fault tolerance to phase discontinuity. Consequently, the reconstructed motions by phase pattern based active stereo vision might have many black holes that are caused by the phase discontinuities. The dot pattern is used to establish the stereo vision correspondences in [8-11]. The advantages of the dot pattern based active stereo vision are its high resistance to external light interference and its high accuracy. The disadvantage of the dot pattern based active stereo vision is its low resolution. Similar to the dot pattern, the advantages of the stripe pattern based active stereo vision [6, 12,13] are its high resistance to external light interference and its high accuracy. The disadvantage of the stripe pattern based active stereo vision is its difficulty to establish exact correspondences between two camera views. Besides these popular structured light patterns, there are also other novel patterns designed for active stereo vision 3D imaging. For instance, the GOBO pattern is generated from the defocused mask pattern for active stereo vision 3D imaging in [14].

In this paper, we also designed a novel RGB dot pattern for active stereo vision because only the dot pattern could establish the exact correspondences between two camera views. The designed pattern contains red dots, green dots and blue dots that form different blocks that will facilitate the stereo vision matching process. The dots in different colors are matched iteratively to rectify errors caused by the occasional imaging differences in two camera views. Although the most used three-dimensional reconstruction principle of stereo vision has remained as disparity for many decades, its validity is not testified. In this paper, we will show that the ray intersection is more robust than disparity in computing the depth for stereo vision technique. As reviewed in [15], both active stereo vision and structured light could achieve high reconstruction accuracy. However, little research work has been conducted to compare their accuracies with the same adopted hardware and the same tested objects. In paper, we will compare them objectively and show that the proposed approach is more robust than existing structured light methods when the adopted projector and cameras are completely the same.

## 2. The Proposed Approach

### 2.1. The 3D imaging system

Figure 1 shows the imaging system of the proposed approach, where a projector projects the designed dot pattern onto the object and two cameras acquire the pattern images in real time. Figure 2a shows our designed pattern that consists of three types of dots in red, green and blue respectively. Specifically, the red dots form the largest block, the green dots form the second largest block and the blue dots form the smallest block. Figure 2b shows the acquired pattern image by the left camera and Figure 2c shows the acquired pattern image by the right camera. We use a hand as the reconstruction target in this paper because it is easy to generate complex 3D motions by a hand.

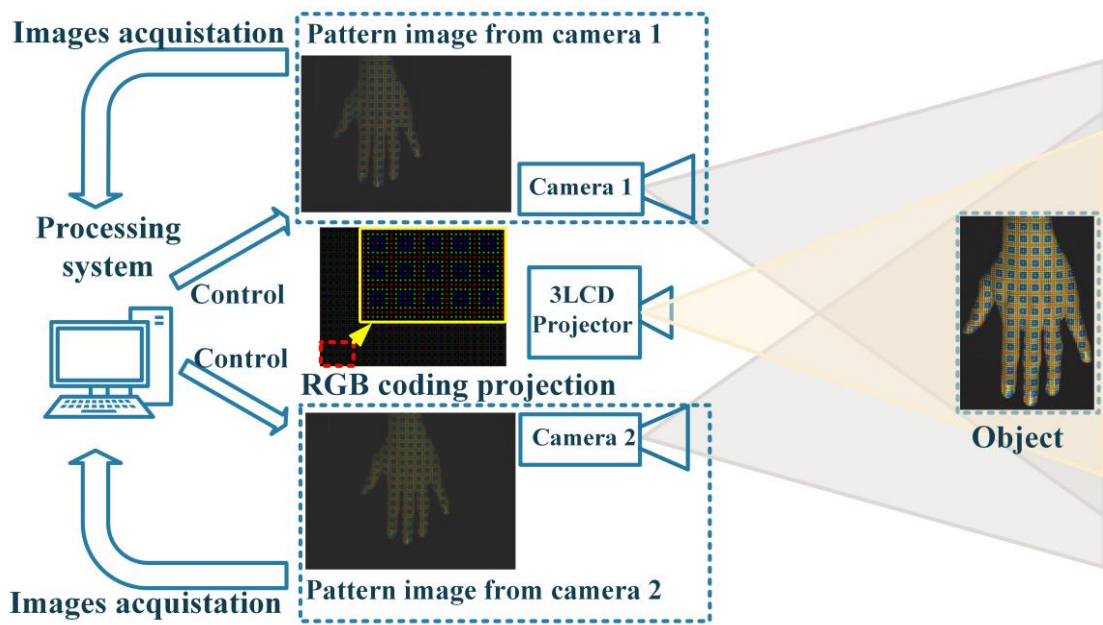

**Figure 1.** The imaging system of the proposed approach.

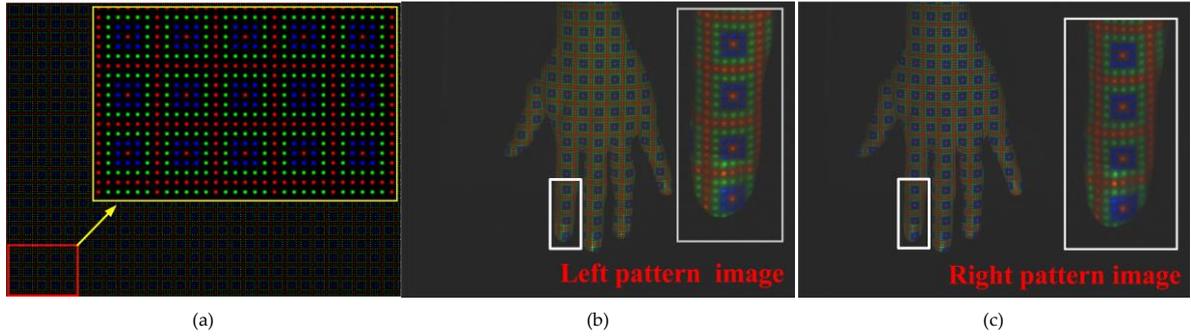

**Figure 2.** Demonstration of the designed pattern and the acquired patterns by two cameras. **(a)**The designed pattern; **(b)** The acquired pattern by the left camera; **(c)** The acquired pattern by the right camera.

With two acquired pattern images, the proposed approach reconstruct the 3D shapes of the object by one-shot. Figure 3 shows the flowchart of the proposed approach. Firstly, the patterns imaged in two cameras are extracted by the proposed image processing method. Secondly, the points in the two patterns are matched by the proposed stereo matching method. Thirdly, the 3D shape of the object is reconstructed by computing the intersection point of two matched 3D lights that are computed by a pair of matched points and the optical centers of the cameras.

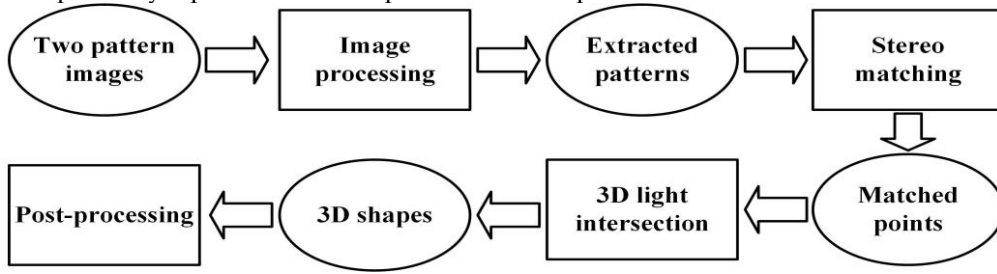

**Figure 3.** The flowchart of the proposed approach.

*2.2. The pattern extraction method*

As demonstrated in Figure 2, the pattern is designed logically. Firstly, the red dots form the connected blocks while the green dots and the blue dots form the non-connected blocks. Our purpose of designing such a pattern is to extract the blocks in different colors and match the corresponding blocks in two camera views. In the RGB color space, the blocks in different color channels are represented by intensities which are easily affected by the interference of external light. From the HSV color model [16], we know that the red blocks, the green blocks and the blue blocks are represented by different hues and could be differentiated easily. Therefore, the acquired pattern image is transformed from the RGB color space to the HSV color space for better pattern extraction accuracy.

In the HSV space, we use the pattern image in the S channel to extract a pattern ROI to constrain the pattern image in the H channel and in the V channel respectively. Figure 4a shows the pattern image in the S channel and Figure 4b shows the segmented ROI by slope different distribution (SDD) threshold selection [17]. Compared to the previous threshold selection methods, the advantages of the SDD threshold selection method include: (1) it could accurately calculate the clustering centers and the partitioning points of the pixel classes in the image histogram by calibrating the two parameters, the bandwidth W of the low-pass discrete Fourier transform (DFT) filter and the fitting number N. (2), the threshold selection rule could be designed specifically and robustly for different applications (3), the thresholds and the clustering centers of the image histogram could always be calculated robustly for any type of images by the following equation.

$$\frac{ds(x)}{dx} = 0 \qquad (1)$$

where $s(x)$ is the slope difference distribution of the image histogram. Figure 4c demonstrates the threshold selection process of SDD. After the threshold $T_s$ is determined by SDD, the pattern in the S channel is segmented as follows.

$$R(i,j) = \begin{cases} 1, & S(i,j) \geq T_s \\ 0, & else \end{cases} \quad (2)$$

where $(i, j)$ denotes the index of the pixel in the image, $R$ denotes the generated ROI image as demonstrated in Figures 4b, and $S$ denotes the pattern image of the S channel respectively.

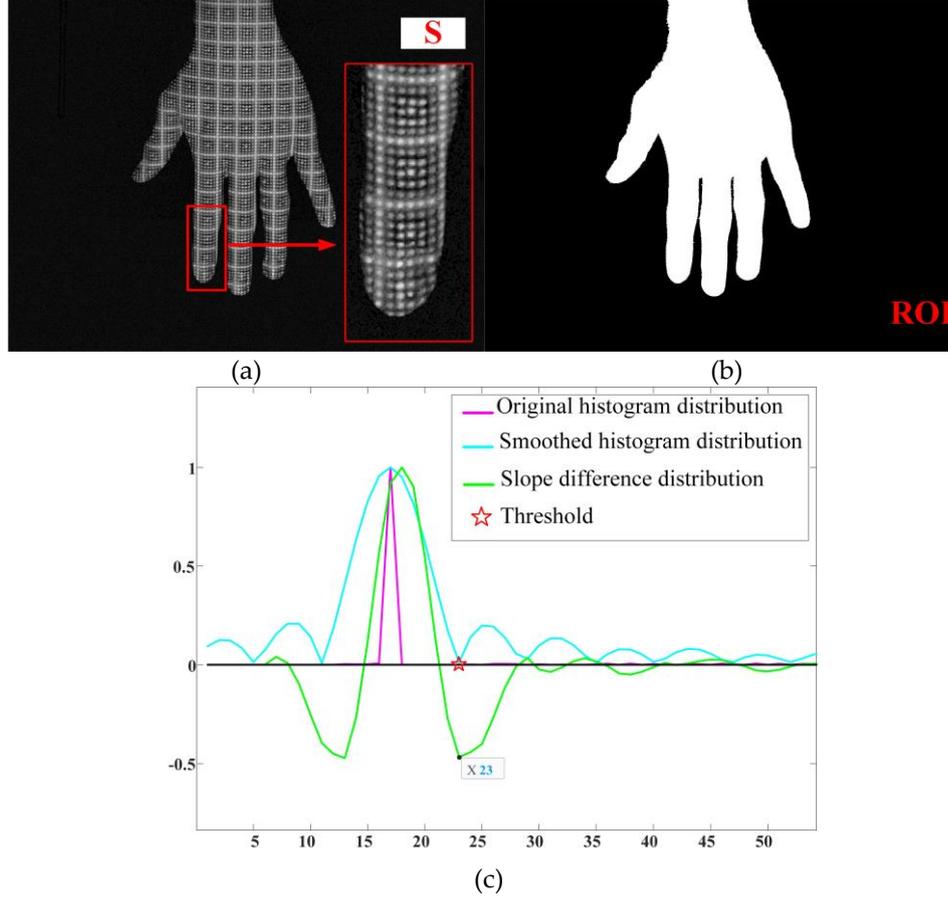

**Figure 4.** Demonstration of segmenting the pattern ROI by SDD. **(a)** The acquired pattern image in the S channel; **(b)** The segmented pattern ROI; **(c)** The threshold selection process by SDD.

In the HSV space, we use the pattern image in the H channel to obtain the red blocks, the green blocks and the blue blocks by SDD clustering segmentation [18]. Figure 5a shows the pattern image in the H channel. From the zoomed in region, we see that the blocks in different colors could be distinguished easily in the H channel. With the constraint of the pattern ROI, the pattern image in the H channel is also segmented by SDD with double thresholds. The calculation of the double thresholds for the red blocks, the green blocks and the blue blocks in the H channel is demonstrated in Figure 5e, where the intensity corresponds to the red circle is the low threshold $T_l^r$ for the red blocks, the intensity corresponds to the red circle is the high threshold $T_h^r$ for the red blocks, the intensity corresponds to the green circle is the low threshold $T_l^g$ for the green blocks, the intensity corresponds to the green star is the high threshold $T_h^g$ for the green blocks, the intensity corresponds to the blue circle is the low threshold $T_l^b$ for the blue blocks, and the intensity corresponds to the blue star is the high threshold $T_h^b$ for the blue blocks. After the thresholds are determined by SDD, the pattern image in the H channel is segmented as follows.

$$C_r(i,j) = \begin{cases} 1, & T_l^r \leq H(i,j) \leq T_h^r \ \& \ R(i,j) > 0 \\ 0, & else \end{cases} \quad (3)$$

$$C_g(i,j) = \begin{cases} 1, T_l^g \le H(i,j) \le T_h^g \ \& \ R(i,j) > 0 \\ 0, else \end{cases} \quad (4)$$

$$C_b(i,j) = \begin{cases} 1, T_l^b \le H(i,j) \le T_h^b \ \& \ R(i,j) > 0 \\ 0, else \end{cases} \quad (5)$$

where $H$ denotes the pattern image of the H channel, $C_r$ denotes the segmentation result of the red blocks, $C_g$ denotes the segmentation result of the green blocks, and $C_b$ denotes the segmentation result of the blue blocks respectively. Figure 5b shows the segmentation result of the red blocks, $C_r$, Figure 5c shows the segmentation result of the green blocks, $C_g$, and Figure 5d shows the segmentation result of the blue blocks, $C_b$ respectively. As can be seen, the segmentation results are satisfactory.

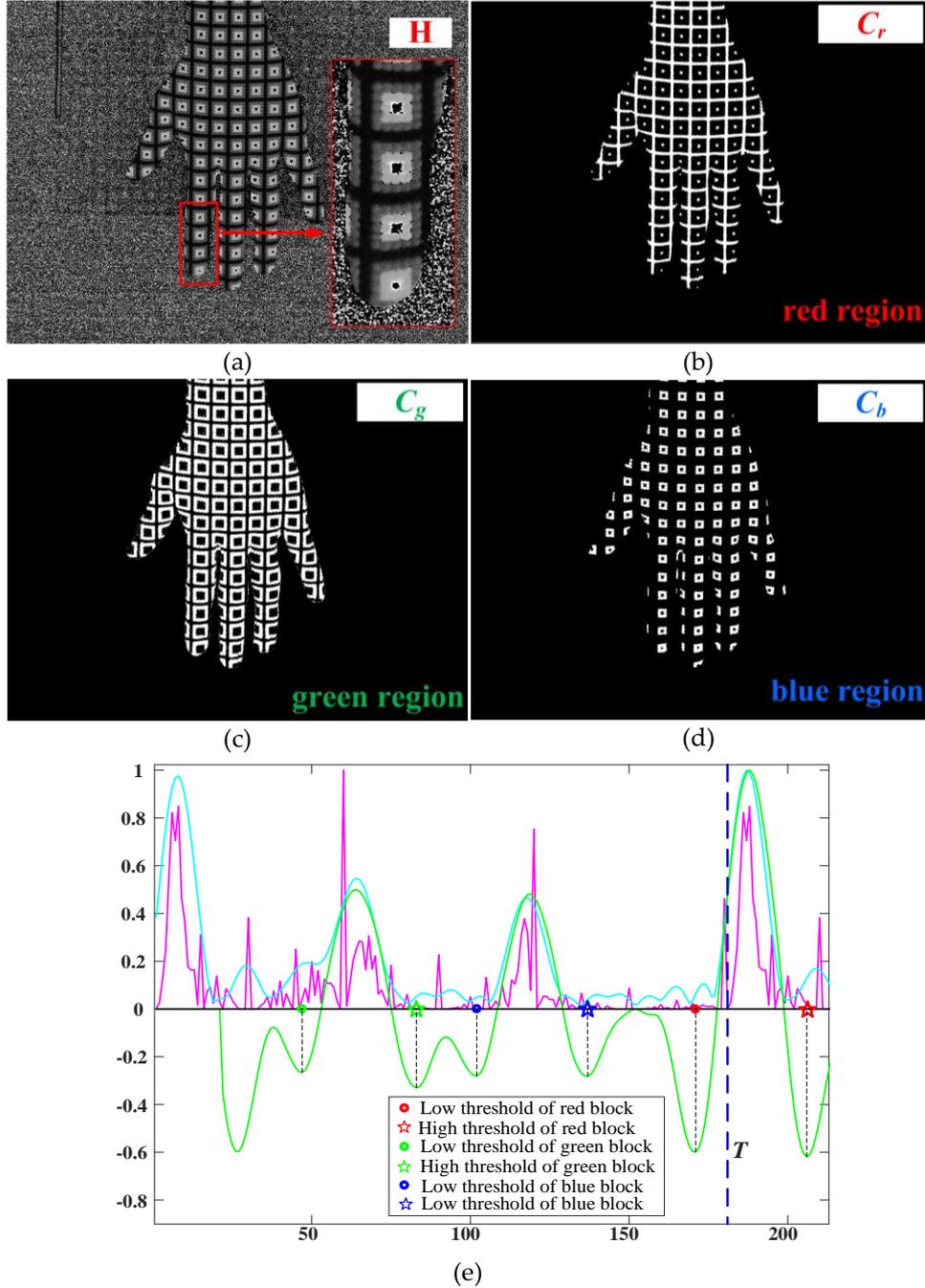

**Figure 5.** Demonstration of segmenting the blocks in different colors in the H channel. **(a)** The acquired pattern image in the H channel; **(b)** The segmented red blocks; **(c)** The segmented green blocks; **(d)** The segmented blue blocks; **(e)** The threshold selection process by SDD.

In the HSV space, we use the pattern image in the V channel to obtain the dots in different colors by detecting the regional maximal intensity values. Figure 6a shows the pattern image in the V channel and Figure 6b shows the dot image $D$ with the detected dots by regional maximal values (RMV). As can be seen, the detected dots contain severe noise. Thus, we need to filter the pattern image before detecting the regional maximal intensity values. Figure 6c shows the dot image $D$ with the detected dots by RMV from the image filtered by a convolution filter defined as follows.

$$V' = V \otimes K \tag{6}$$

Where $V$ denotes the pattern image in the $V$ channel and $V'$ denotes the filtered pattern image in the V channel. $K$ denotes the convolution kernel and it is a 5×5 matrix with all values equal to 1/25. As can be seen, the noises have been removed successfully.

Figure 6d shows the dot image $D$ with the detected dots by RMV from the image filtered by a morphological opening filter defined as follows.

$$V' = V \circ B = (I \ominus B) \oplus B \tag{7}$$

where $B$ denotes a disk structuring element with the radius equal to 1, i.e. $B = \{(0,0)\}$. As can be seen, the noises are also been removed successfully. The difference between the detected dots with convolution filter and the detected dots with morphological opening filter is that the dots with convolution filter have only one pixel while the dots with morphological filter have multiple pixels. To compute the positions of the dots with sub-pixel accuracy, we choose the morphological opening filter in this paper.

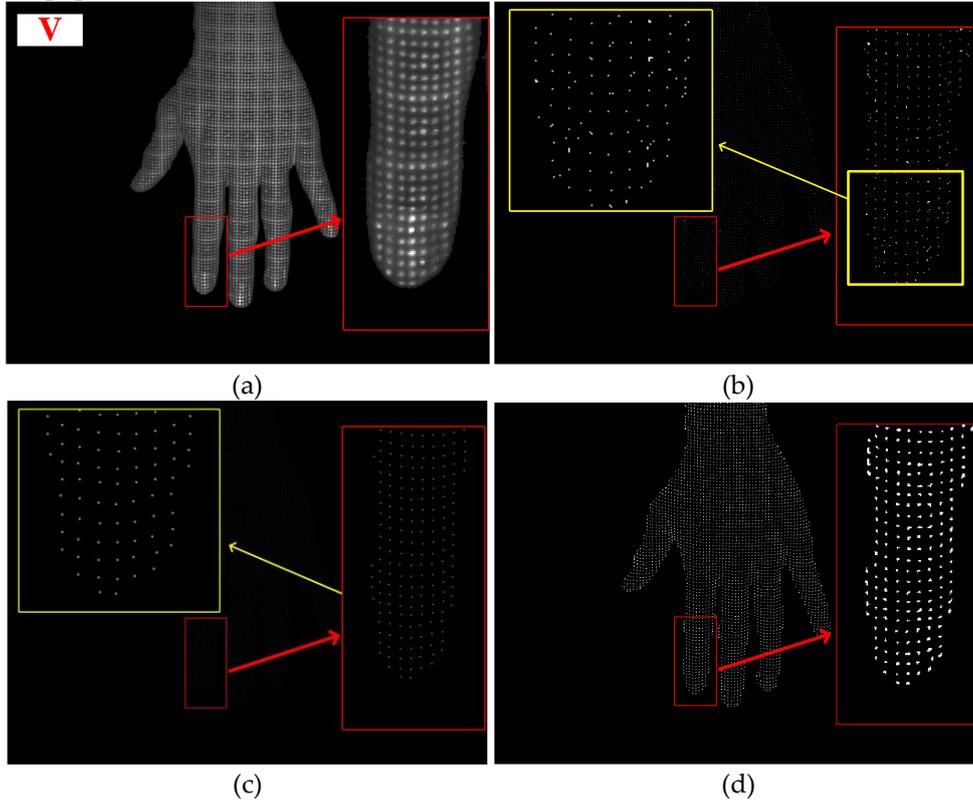

**Figure 6.** Demonstration of extracting the dots in different colors in the V channel. **(a)** The acquired pattern image in the V channel; **(b)** The extracted dots by detecting the regional maximal values without filtering; **(c)** The extracted dots by detecting the regional maximal values with a convolution filter; **(d)** The extracted dots by detecting the regional maximal values with a morphological opening filter.

The dots in different colors are computed as follows.

$$D_r(i,j) = \begin{cases} 1, & C_r(i,j) > 0 \ \& \ D(i,j) > 0 \\ 0, & else \end{cases} \tag{8}$$

$$D_g(i,j) = \begin{cases} 1, C_g(i,j) > 0 \,\&\, D(i,j) > 0 \\ 0, else \end{cases} \quad (9)$$

$$D_b(i,j) = \begin{cases} 1, C_b(i,j) > 0 \,\&\, D(i,j) > 0 \\ 0, else \end{cases} \quad (10)$$

where $D_r$ notes the dot image that contains the detected red dots, $D_g$ notes the dot image that contains the detected green dots, and $D_b$ notes the dot image that contains the detected blue dots respectively. The position $(x_k, y_k)$ of the $k$th detected dot in the dot image are computed as follows.

$$(x_k, y_k) = \left( \frac{1}{M} \sum_{m=1}^{M} x_m^k, \frac{1}{M} \sum_{m=1}^{M} y_m^k \right) \quad (11)$$

where $(x_m^k, y_m^k), m = 1, 2, ..., M$ is the position of $m$th pixel of the dot in the pattern image and $M$ is the total number of pixel contained in the $k$th detected dot. Figure 7a plots the computed positions for the red dots, Figure 7b plots the computed positions for the green dots and Figure 7c plots the computed positions for the blue dots respectively. Figure 7d plots the computed positions for all the dots in red, green and blue. As can be seen, the computed positions of the dots have high correlation with the positions of the imaged dots as shown in Figure 2.

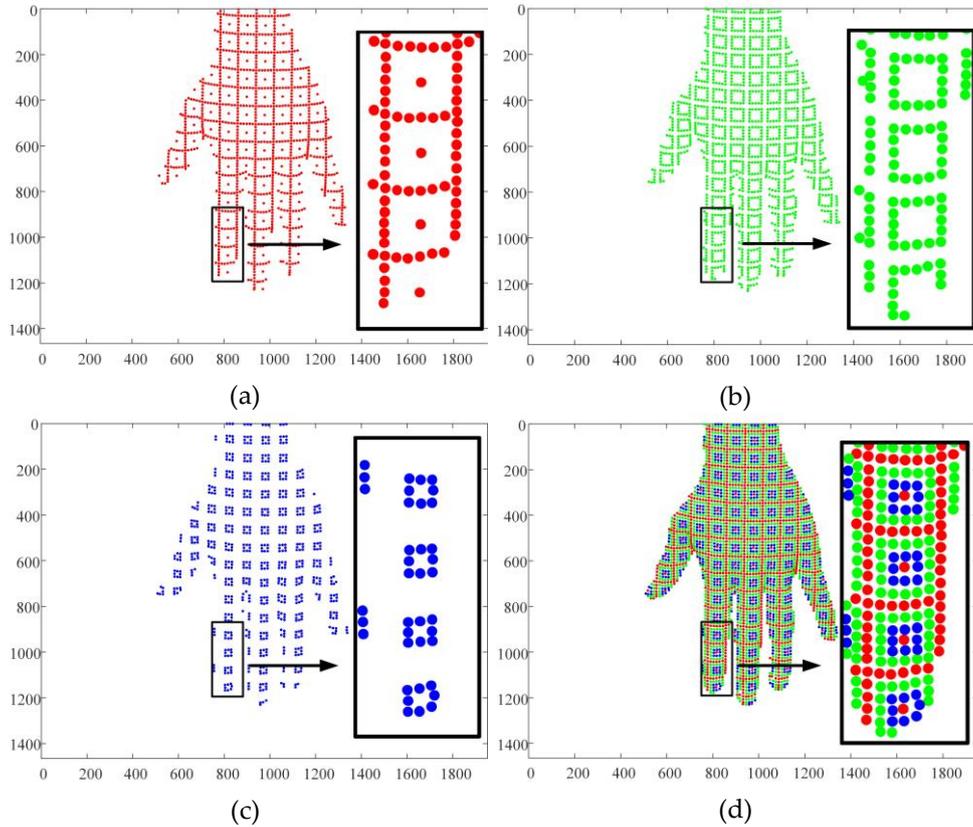

**Figure 7.** Demonstration of the computed positions for the dots in different colors. (a)The computed positions for the dots in red; (b) The computed positions for the dots in green; (c) The computed positions for the dots in blue; (d) The computed positions for the dots in red, green and blue.

*2.3. The 3D imaging system*

Before matching the dots in the two patterns from two camera views, we align the two patterns by matching the two ROIs that are computed from the S channels of the two pattern images (demonstrated in Figure 4b). The ROI, $R_l$ computed from the left pattern image and the ROI $R_r$ computed from the right pattern image are matched as follows.

**Step 1:** the center of the left ROI, $R_l$ is computed as $(x_c^l, y_c^l)$ and the center of the right ROI, $R_r$ is computed as $(x_c^r, y_c^r)$ respectively.

**Step 2:** shift the left ROI to the left in the horizontal direction with the distance $|x_c^r - x_c^l|$ and align the two ROIs roughly.

**Step 3:** shift the left ROI in the horizontal direction with $\Delta$ in the range [-10, 10] and find $\Delta$ that makes the overlapping area of the two ROIs maximum:

$$\Delta' = \max_{\Delta \in [-10,10]} R_l \cap R_r \quad (12)$$

**Step 4:** Finally, the two ROIs are aligned by shifting the left ROI to the left in the horizontal direction with the distance $d = |x_c^r - x_c^l| + \Delta'$.

The green blocks in the segmented pattern image $C_g$ and the blue blocks in the segmented pattern image $C_b$ are matched by the following proposed block matching method respectively.

**Step 1:** shift the blocks in the left pattern image to the left in the horizontal direction with the distance $d = |x_c^r - x_c^l| + \Delta$ to align the blocks from the pair of pattern images. Figure 8a demonstrates the aligned green blocks from the pair of pattern images and Figure 8b demonstrates the aligned blue blocks from the pair of pattern images respectively. As can be seen, all the blocks are aligned correctly.

**Step 2:** label the blocks in the left segmented pattern image ($C_g$ or $C_b$) to get the labeled image, $L^L$ and the label numbers of the blocks, $L_{n_l}^L, n_l = 1, 2, ..., N^L$. Label the blocks in the right segmented pattern image ($C_g$ or $C_b$) to get the labeled image $L^R$ and the label numbers of the blocks, $L_{n_r}^L, n_r = 1, 2, ..., N^R$. Update the left labeled image by the following equation.

$$L^{L'}(i, j) = L^L(i, j) * 10^\rho \quad (13)$$

$$\rho = 1 + \lfloor \log(N^R) \rfloor \quad (14)$$

**Step 3:** Add the updated left image and the right labeled image to get the overlapping labeled image, $L^O$.

$$L^O = L^{L'} + L^R \quad (15)$$

The label number of the *n*th overlapped block in the overlapping labeled image is computed as:

$$L_n^O = L_{n_l}^L * 10^\rho + L_{n_r}^R \quad (16)$$

where $n_l$ notes the $n_l$th labeled block in the left labeled image, $L^L$ and $n_r$ denotes the $n_r$th labeled block in the right labeled image, $L^R$. Accordingly, the first $\rho$ digits of $L_n^O$ denotes the label number of the $n_l$th labeled block in the left labeled image and the last $\rho$ digits of $L_n^O$ denotes the label number of the $n_r$th labeled block in the right labeled image. As a result, the labeled blocks in the left image and the labeled blocks in the right image are matched based on the label number of $L_n^O$. Figure 9 demonstrates the computed matching relationships between the labeled green blocks from two pattern images.

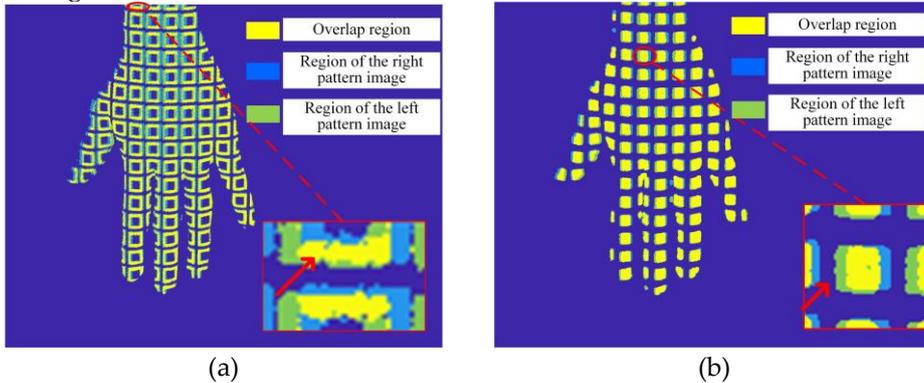

**Figure 8.** Demonstration of the aligned blocks by shifting; **(a)** the aligned green blocks from the pair of pattern images; **(b)** the aligned blue blocks from the pair of pattern images.

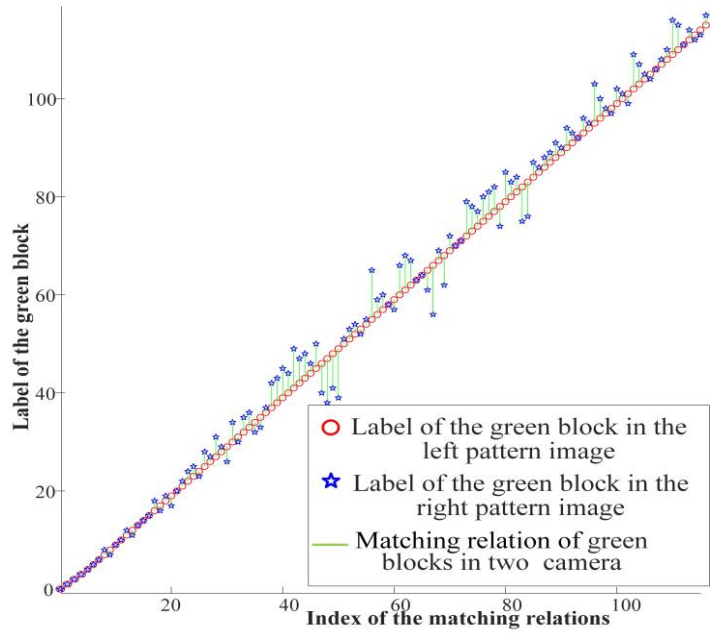

**Figure 9.** Demonstration of the matching relationships between the green blocks in two pattern images.

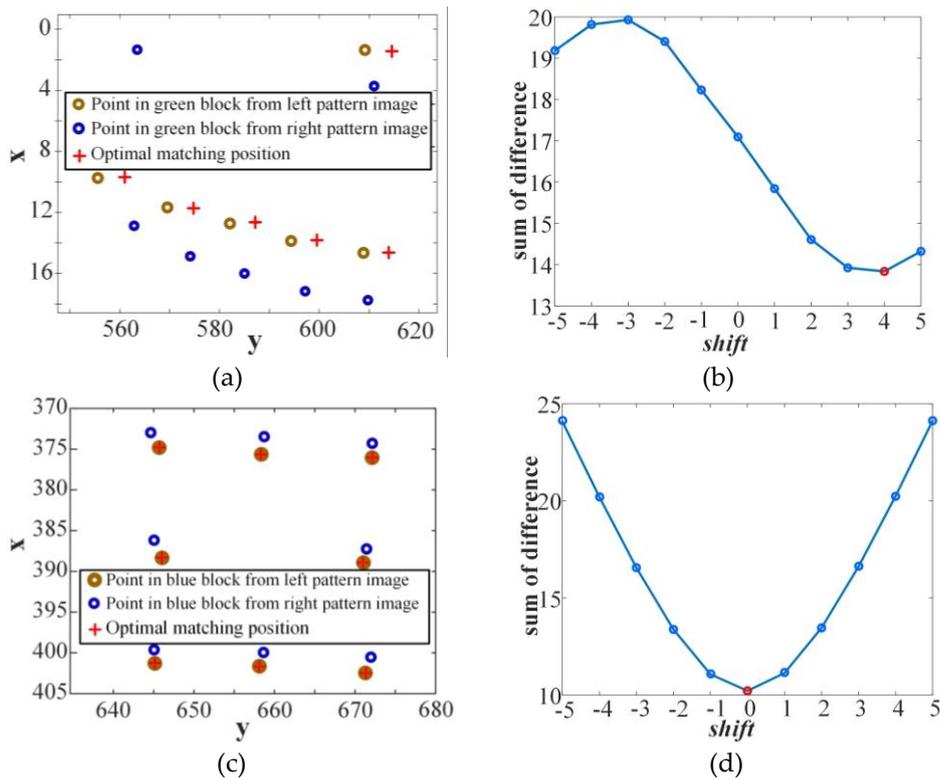

**Figure 10.** Demonstration of matching the dots of zoomed-in blocks in Figure 8; **(a)** the final matching positions of the dots in the zoomed-in green block; **(b)** the computed sum of differences with different shifted values for the demonstrated green block; **(c)** the final matching positions of the dots in the zoomed-in blue block; **(d)** the computed sum of differences with different shifted values for the demonstrated blue block.

After the green blocks and the blue blocks in the pair of segmented pattern images $C_g$ are matched, the dots contained in the matched blocks are matched by the following method.

**Step 1:** shift the dots in the left matched block with $\delta$ in the range [-5, 5] and compute the sum of the differences between the dots by the following equation.

$$S_D = \sum_{m1=1}^{M1} \sum_{m2=1}^{M2} \sqrt{\left(x_{m1}^L - \delta - x_{m2}^R\right)^2 + \left(y_{m1}^L - y_{m2}^R\right)^2} \tag{17}$$

where $\left(x_{m1}^L, y_{m1}^L\right)$ denotes the coordinate of the $m$1th dot in the left matched block and $\left(x_{m2}^R, y_{m2}^R\right)$ denotes the coordinate of the $m$2th dot in the right matched block. $M1$ denotes the total number of dots in the left matched block and $M2$ denotes the total number of dots in the right matched block.

**Step 2:** select the value of $\delta'$ that makes the sum of differences, $S_D$ minimum and shift the dots in the left matched block with $\delta'$.

**Step 3:** Find the best matched dot of each dot $\left(x_{m1}^L - \delta', y_{m1}^L\right)$ in the shifted left matched block from the right matched block by making the following equation minimum.

$$m' = \min_{m \in [1, M2]} \sqrt{\left(x_{m1}^L - \delta' - x_m^R\right)^2 + \left(y_{m1}^L - y_m^R\right)^2} \tag{18}$$

$\left(x_{m'}^R - \delta', y_{m'}^R\right)$ is the coordinate of the matched dot in the right matched block. Figure 10 demonstrates the process of matching the dots of the zoomed-in blocks in Figure 8. In Figure 10a and c, the brown circles denote the original positions of the dots in the left matched block and the blue circles denote the original positions of the dots in the right matched block. The red crosses denote the new positions of the dots in the left matched block after shifting $\delta'$ in the horizontal direction. From Figure 10b and d, it is seen that the shifting value $\delta' = 4$ and $\delta' = 0$ respectively.

After the dots contained in the green blocks and the blue blocks are matched between the left pattern image and the right pattern image, the mapping function $\left(f_x^{gb}, f_y^{gb}\right)$ between the green & blue dots in the left pattern image and the green & blue dots in the right pattern image are calculated by interpolating scattered data [19] as follows.

$$\left(x_i^{Rgb}, y_i^{Rgb}\right) = \left(f_x^{gb}\left(x_i^{Lgb}, y_i^{Lgb}\right), f_y^{gb}\left(x_i^{Lgb}, y_i^{Lgb}\right)\right) \tag{19}$$

where $\left(x_i^{Lgb}, y_i^{Lgb}\right), i = 1, 2, ..., N_{gb}$ denotes the coordinates of the dots contained in the left green and blue blocks. $\left(x_i^{Rgb}, y_i^{Rgb}\right), i = 1, 2, ..., N_{gb}$ denotes the coordinates of the dots contained in the right green and blue blocks. $N_{gb}$ denotes the total number of matched dots in the green and blue blocks. With the mapping function $\left(f_x^{gb}, f_y^{gb}\right)$, the dots in the right green blocks could be fitted by the dots in the left green blocks as follows.

$$\left(x_i^{Fg}, y_i^{Fg}\right) = \left(f_x^{gb}\left(x_i^{Lg}, y_i^{Lg}\right), f_y^{gb}\left(x_i^{Lg}, y_i^{Lg}\right)\right) \tag{20}$$

where $\left(x_i^{Lg}, y_i^{Lg}\right), i = 1, 2, ..., N_g$ denotes the coordinates of the dots contained in the left green blocks. $\left(x_i^{Fg}, y_i^{Fg}\right), i = 1, 2, ..., N_g$ denotes the coordinates of the fitted green dots. $N_g$ denotes the total number of matched dots in the green blocks.

Similarly, the dots in the right blue blocks could be fitted by the dots in the left blue blocks as follows.

$$\left(x_i^{Fb}, y_i^{Fb}\right) = \left(f_x^{gb}\left(x_i^{Lb}, y_i^{Lb}\right), f_y^{gb}\left(x_i^{Lb}, y_i^{Lb}\right)\right) \tag{21}$$

where $\left(x_i^{Lb}, y_i^{Lb}\right), i = 1, 2, ..., N_b$ denotes the coordinates of the dots contained in the left blue blocks. $\left(x_i^{Fb}, y_i^{Fb}\right), i = 1, 2, ..., N_b$ denotes the coordinates of the fitted blue dots. $N_g$ denotes the total number of matched dots in the green blocks. Figure 11a demonstrates the matching fitted dots $\left(x_i^{Lg}, y_i^{Lg}\right)$ from the left green blocks and the matched actual dot $\left(x_i^{Rg}, y_i^{Rg}\right)$ from the right green blocks. Figure 11b demonstrates the matching fitted dots $\left(x_i^{Lb}, y_i^{Lb}\right)$ from the left blue blocks and the matched actual dots $\left(x_i^{Rb}, y_i^{Rb}\right)$ from the right blue blocks. As can be been, there are many unmatched dots, especially in the zoomed-in region of Figure 11a.

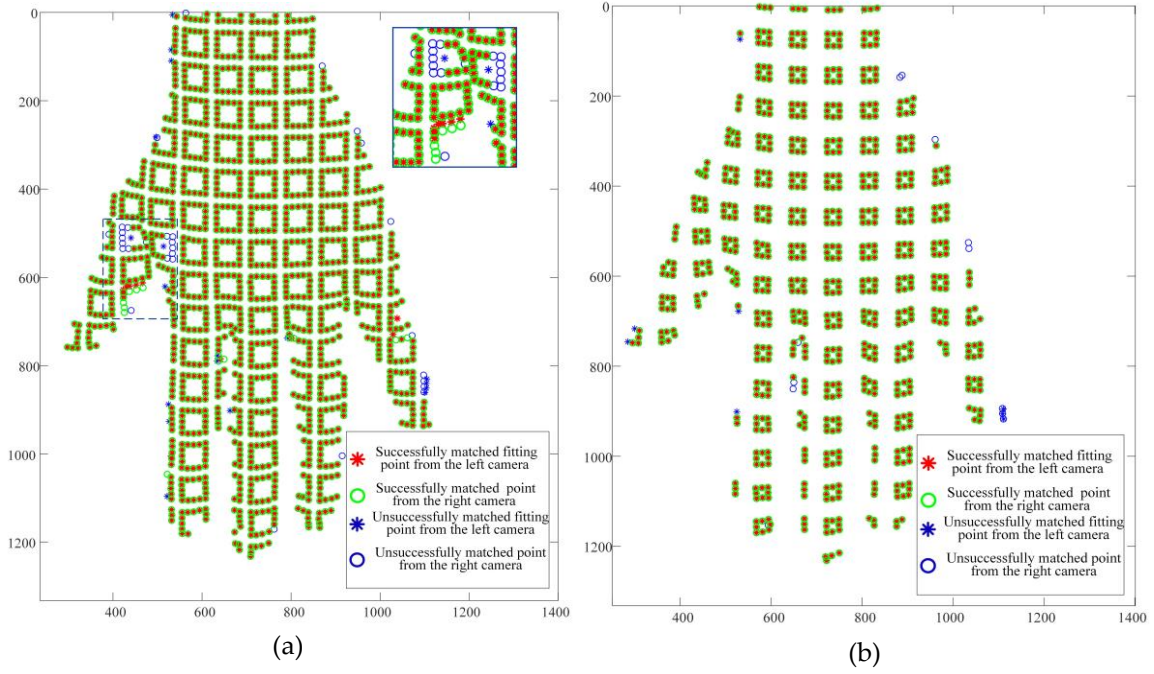

**Figure 11.** Demonstration of the matching fitted dots from the left blocks and the matching actual dots from the right blocks. **(a)** the matching dots from the green blocks; **(b)** the matching dots from the blue blocks.

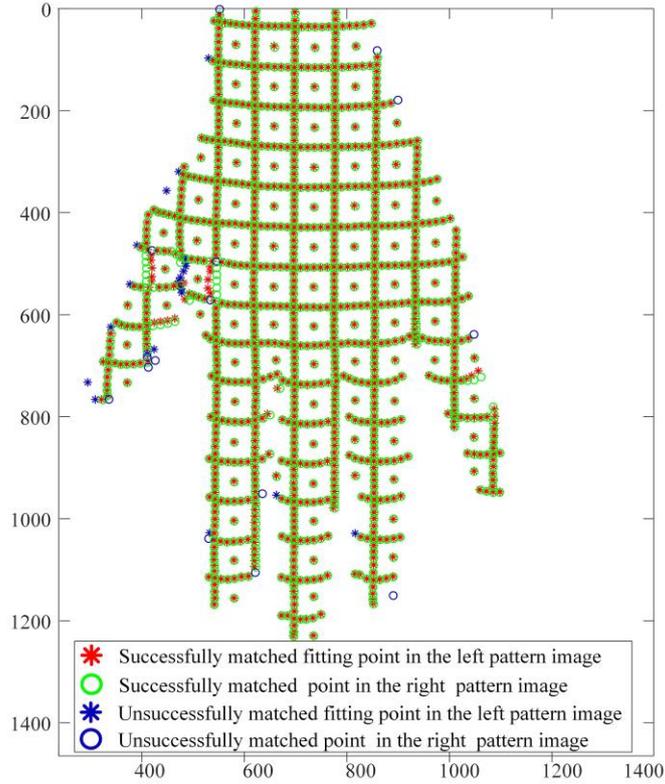

**Figure 12.** Demonstration of the matching fitted dots from the left red blocks and the matching actual dots from the right red blocks.

With the same mapping function $\left(f_x^{gb}, f_y^{gb}\right)$, the dots in the right red blocks could be fitted by the dots in the left red blocks as follows.

$$\left(x_i^{Fr}, y_i^{Fr}\right) = \left(f_x^{gb}\left(x_i^{Lr}, y_i^{Lr}\right), f_y^{gb}\left(x_i^{Lr}, y_i^{Lr}\right)\right) \tag{22}$$

Where $(x_i^{Lr}, y_i^{Lr}), i = 1, 2, ..., N_r$ denotes the coordinates of the dots contained in the left red blocks. $(x_i^{Fr}, y_i^{Fr}), i = 1, 2, ..., N_r$ denotes the coordinates of the fitted red dots. $N_r$ denotes the total number of dots in the left red blocks. As can be seen, the dots denoted by $(x_i^{Lr}, y_i^{Lr})$ and their fitted dots denoted by $(x_i^{Fr}, y_i^{Fr})$ are matched one by one based on their index $i$. To match the dots $(x_i^{Lr}, y_i^{Lr})$ in the left red blocks and the dots $(x_i^{Rr}, y_i^{Rr})$ in the right red blocks, we just need to match the fitted dots $(x_i^{Fr}, y_i^{Fr})$ and the dots $(x_i^{Rr}, y_i^{Rr})$ in the right red blocks. The $j$th fitted dot $(x_j^{Fr}, y_j^{Fr}), j = 1, 2, ..., N_r$ by the dots in the left red blocks and the matched $i'$th actual dot $(x_{i'}^{Rr}, y_{i'}^{Rr})$ in the right red blocks are matched by the following equation.

$$i' = \min_{i \in [1, N_r]} \sqrt{\left(x_j^{Fr} - x_i^{Fr}\right)^2 + \left(y_j^{Fr} - y_i^{Fr}\right)^2} \tag{23}$$

Figure 12 demonstrates the matching fitted dots $(x_i^{Fr}, y_i^{Fr})$ from the left red blocks and the matched actual dots $(x_i^{Rr}, y_i^{Rr})$ from the right red blocks. As can be been, there are many unmatched dots, especially in the zoomed-in region of Figure 11a. To match all the dots successfully, we propose the following iterative dot matching method.

**Step 1:** calculate the mapping function $(f_x^{gb}, f_y^{gb})$ between the matched green & blue dots in the left pattern image and the matched green & blue dots in the right pattern image by Equation (24). Calculate the fitted red dots by Equation (25) and find their matched actual red dots by Equation (26). The matched red dots in the left red blocks and in the right red blocks are updated.

**Step 2:** calculate the mapping function $(f_x^{rg}, f_y^{rg})$ between the matched red & green dots in the left pattern image and the matched red & green dots in the right pattern image by the following equation.

$$\left(x_i^{Rrg}, y_i^{Rrg}\right) = \left(f_x^{rg}\left(x_i^{Lrg}, y_i^{Lrg}\right), f_y^{rg}\left(x_i^{Lrg}, y_i^{Lrg}\right)\right) \tag{27}$$

Fit the dots in the right blue blocks by the following equation.

$$\left(x_i^{Fb}, y_i^{Fb}\right) = \left(f_x^{rg}\left(x_i^{Lb}, y_i^{Lb}\right), f_y^{rg}\left(x_i^{Lb}, y_i^{Lb}\right)\right) \tag{28}$$

Find the actual matched blue dots in the right blue blocks by making the following equation minimum.

$$\min_{i \in [1, N_b]} \sqrt{\left(x_j^{Fb} - x_i^{Rb}\right)^2 + \left(y_j^{Fb} - y_i^{Rb}\right)^2} \tag{29}$$

The matched blue dots in the left blue blocks and in the right blue blocks are updated.

**Step 3:** calculate the mapping function $(f_x^{rb}, f_y^{rb})$ between the matched red & blue dots in the left pattern image and the matched red & blue dots in the right pattern image by the following equation.

$$\left(x_i^{Rrb}, y_i^{Rrb}\right) = \left(f_x^{rb}\left(x_i^{Lrb}, y_i^{Lrb}\right), f_y^{rb}\left(x_i^{Lrb}, y_i^{Lrb}\right)\right) \tag{30}$$

Fit the dots in the right green blocks by the following equation.

$$\left(x_i^{Fg}, y_i^{Fg}\right) = \left(f_x^{rb}\left(x_i^{Lg}, y_i^{Lg}\right), f_y^{rb}\left(x_i^{Lg}, y_i^{Lg}\right)\right) \tag{31}$$

Find the actual matched green dots in the right green blocks by making the following equation minimum.

$$\min_{i \in [1, N_g]} \sqrt{\left(x_j^{Fg} - x_i^{Rg}\right)^2 + \left(y_j^{Fg} - y_i^{Rg}\right)^2} \tag{32}$$

The matched green dots in the left green blocks and in the right green blocks are updated.

**Step 4:** repeat Steps 1-3 until all the matched dots in red, green and blue blocks become unchanged after update.

Figure 13 shows the final matched dots in all blocks. As can be seen, all the paired dots are matched successfully.

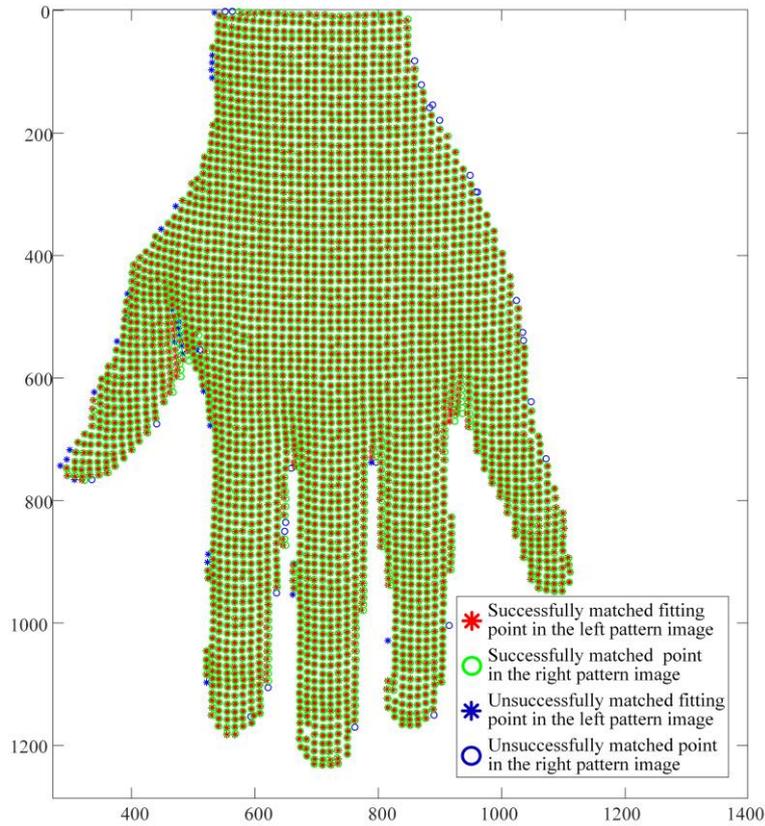

**Figure 13.** Demonstration of the matching fitted dots from the left blocks and the matching actual dots from the right blocks computed by the proposed iterative matching method.

*2.4. Three-dimensional reconstruction*

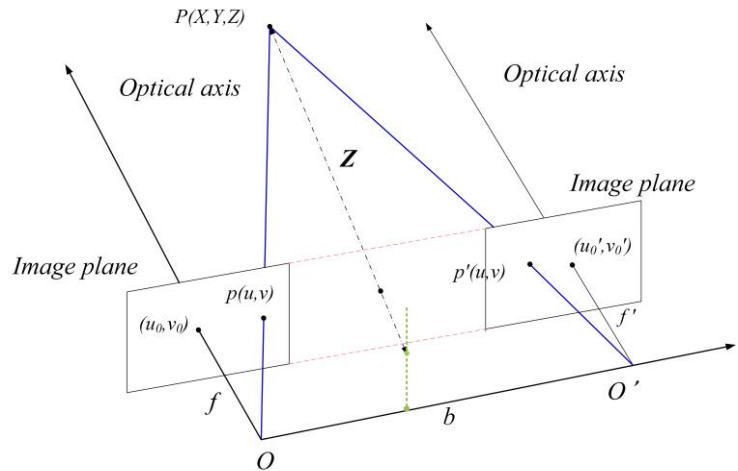

**Figure 14.** Demonstration of the matching fitted dots from the left blocks and the matching actual dots from the right blocks computed by the proposed iterative matching method.

There are two well-known three-dimensional reconstruction methods for stereo vision as described in [20]. One is the disparity method and the other is the ray intersection method. For many decades, the disparity method remained as the most widely used method for stereo vision due to its simplicity [21]. However, little work has been done to compare the accuracies of these two methods. Thus, we implemented both methods to compare their accuracies in our experiment. The ray intersection method was called middle-point algorithm in [20] and its implementation is based on calculating the shortest line segment between the two corresponding rays. The ray intersection method derived in [8-10] is named as analytical solution algorithm and it is illustrated in Figure 14.

The equation of the image plane in the left camera is calculated as $\pi = [0,0,-1,f]$ and the equation of the image plane in the right camera is calculated as $\pi = [0,0,-1,f']$. The equation of the light $\overrightarrow{OpP}$ is thus calculated as:

$$\frac{x-O(1)}{O(1)-p(1)} = \frac{y-O(2)}{O(2)-p(2)} = \frac{z-O(3)}{O(3)-f} = t \tag{33}$$

The equation of the light $\overrightarrow{O'p'P}$ is thus calculated as:

$$\frac{x-O'(1)}{O'(1)-p(1)} = \frac{y-O'(2)}{O'(2)-p(2)} = \frac{z-O'(3)}{O'(3)-f} = t \tag{34}$$

where $f$, $f'$, $O$ and $O'$ are obtained by stereo vision calibration by Zhang's method [22]. $p$ and $p'$ are the pixel coordinate of the left camera and the pixel coordinate of the right camera respectively. The 3D coordinate of the point is calculated as the intersection points of the light $\overrightarrow{OpP}$ and the light $\overrightarrow{O'p'P}$ by the method proposed in [8-10].

After the 3D shape of the object is reconstructed based on all the matched points as demonstrated in Figure 13, the resolution of the reconstructed 3D shape is increased by interpolating scattered data [19] as follows.

**Step 1:** the mapping function $f^r$ between the depth $Z_i^r$ of the reconstructed shape and the 2D coordinate $(X_i^r, Y_i^r), i = 1,2,...N^r$ is calculated by the following equation.

$$Z_i^r = f^r(X_i^r, Y_i^r) \tag{35}$$

**Step 2:** the 2D coordinate $(X_i^r, Y_i^r)$ is replaced with a 2D grid $(X_i^{r'}, Y_i^{r'}), i = 1,2,...N^S$ with uniform sampling and the corresponding depth $Z_j^{r'}$ is computed as:

$$Z_j^{r'} = f^r(X_j^{r'}, Y_j^{r'}) \tag{36}$$

where $N^r$ is the total number of the reconstructed points and $N^S$ is the total number of the sampled points. Figure 15 demonstrates the reconstructed points overlaying on the interpolated 3D surface.

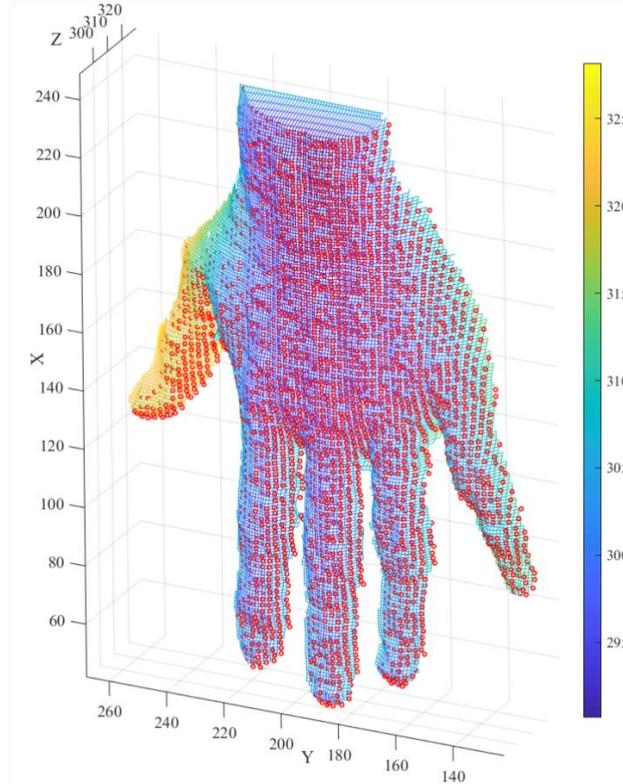

**Figure 15.** Demonstration of the interpolated 3D surface with the original reconstructed points (denoted by red circles) overlaying on it.

## 3. Results and Discussion

Figure 16 shows our established imaging system, where a NEC NP-P451X+ 3LCD projector is used to project the designed RGB dot pattern onto the object and two BFLY-PGE-31S4C-C cameras with the resolution of 2048× 1536 are used to capture the images in real time with the frame rate 35f/s. The reconstruction accuracy is evaluated with 25 marker points on a designed grid with known world coordinates as shown in Figure 17. The mean squared error (MSE) between the ground-truth world coordinates and the reconstructed coordinates for the 25 marker points is computed as:

$$MSE = \frac{1}{25}\sum_{i=1}^{25}\left(X_i^r - X_i^W\right)^2 + \left(Y_i^r - Y_i^W\right)^2 + \left(Z_i^r - Z_i^W\right)^2 \qquad (37)$$

where $\left(X_i^r, Y_i^r, Z_i^r\right), i=1,2,\ldots,25$ is the coordinate of the reconstructed marker point on the grid and $\left(X_i^W, Y_i^W, Z_i^W\right), i=1,2,\ldots,25$ is the ground truth coordinate of the marker point on the grid. The calculated mean squared error (MSE) is 0.31 mm.

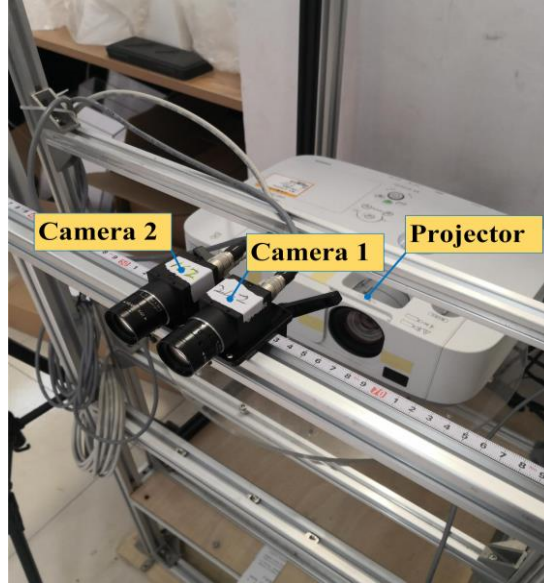

**Figure 16.** Our established system.

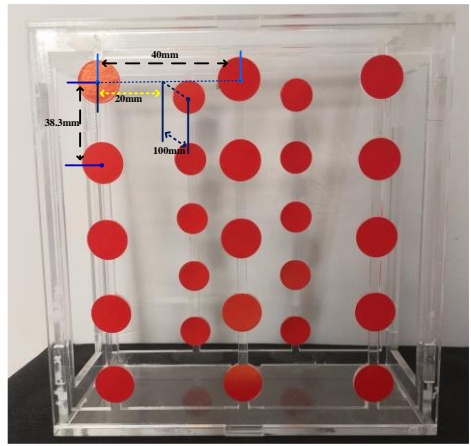

**Figure 17.** Our established system.

Besides the grid, we also use a ball to evaluate the reconstruction accuracy of the proposed approach quantitatively. The ball is shown in Figure 18 a and b and its reconstruction results in different view angles are shown in Figure 18 c and d respectively. The reconstructed points are used to fit the equation of the ball and the radius of the ball is computed as 84.6763 mm. The mean distance (MD) between the fitted surface and the reconstructed points is computed as:

$$MD = \frac{1}{N^b}\sum_{i=1}^{N^b} D_i^b \qquad (38)$$

where $D_i^b$ denotes the distances between the $i$th reconstructed point and the fitted ball surface and $N^b$ denotes the total number of the reconstructed points. The calculated mean distance (MD) is 0.3358 mm. The quantitative comparison of the reconstruction accuracy with state of the art methods is shown in Table 1. The time of flight method is tested with Kinect v2. The compared structured light methods include the one-shot line pattern (LP) method [23] and the multiple-shot multiple frequency phase shifting profilometry (MFPSP) method [24]. Both structured light systems were established with the same camera and projector as those used by the proposed approach for objective comparison. As can be seen, the ray intersection method is more robust than the disparity method. The proposed approach is more accurate than the time of the flight method and the structured light methods [23-24]. Since different algorithms have been used to implement the ray intersection method in [8] and [20], it is necessary to compare their accuracies and efficiencies. Four algorithms for ray intersection computation were compared in [25] and one of them is the middle-point algorithm as described in [20]. The middle-point algorithm was implemented following the steps described in [25] and the quantitative accuracy comparison between it and the analytical solution algorithm [8] is shown in Table 2. As can be seen, the computation time of the analytical solution algorithm is 60 times faster than that of the middle point algorithm while the reconstruction errors did not have noticeable differences.

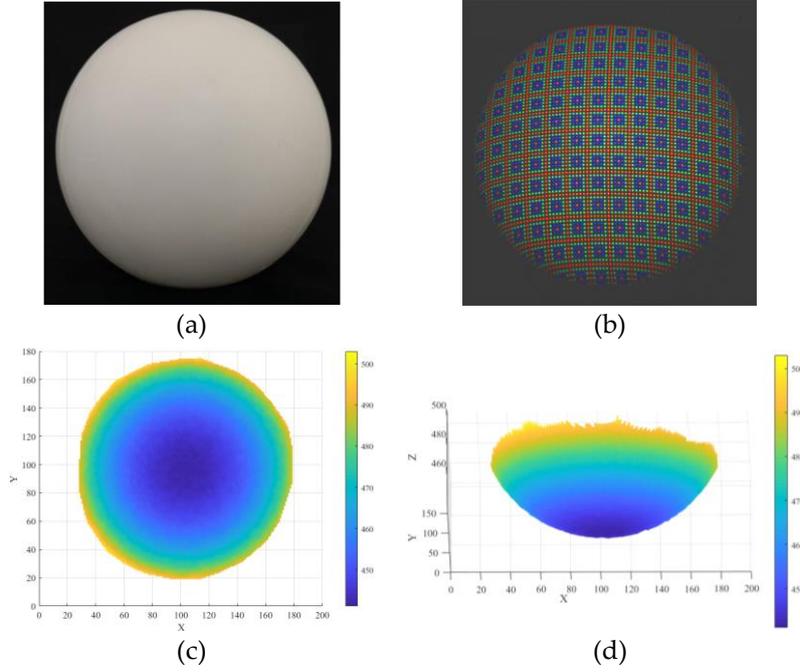

**Figure 18.** Results of reconstructing a ball; **(a)** the ball without structured light pattern; **(b)** the ball with structured light pattern; **(c)** the reconstructed ball in view 1; **(d)** the reconstructed ball in view 2.

Table 1. Comparison of the accuracies of different 3D reconstruction methods

| Methods\Objects | Grid | Ball |
| --- | --- | --- |
| Time of flight | 3.5 mm | 0.92 mm |
| LP [23] | 0.36 mm | 0.342 mm |
| MFPSP [24] | 0.33 mm | 0.339 mm |
| Disparity | 0.42mm | 0.354 mm |
| Ray intersection | 0.31 mm | 0.336 mm |

Table 2. Comparison of the accuracies and efficiencies of different ray intersection algorithms

| Algorithms | Times | Grid | Ball |
|---|---|---|---|
| Middle-point | 0.3 s | 0.31 mm | 0.336 mm |
| Analytic solution | 0.005 s | 0.31 mm | 0.336 mm |

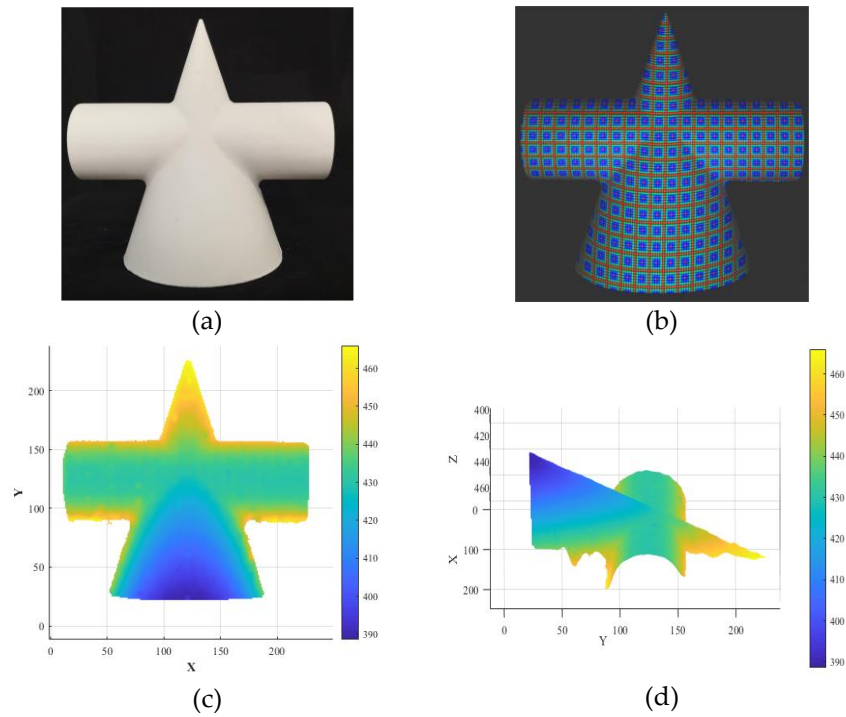

**Figure 19.** Results of reconstructing a cone with cylinder; **(a)** the cone without structured light pattern; **(b)** the cone with structured light pattern; **(c)** the reconstructed cone in view 1; **(d)** the reconstructed cone in view 2.

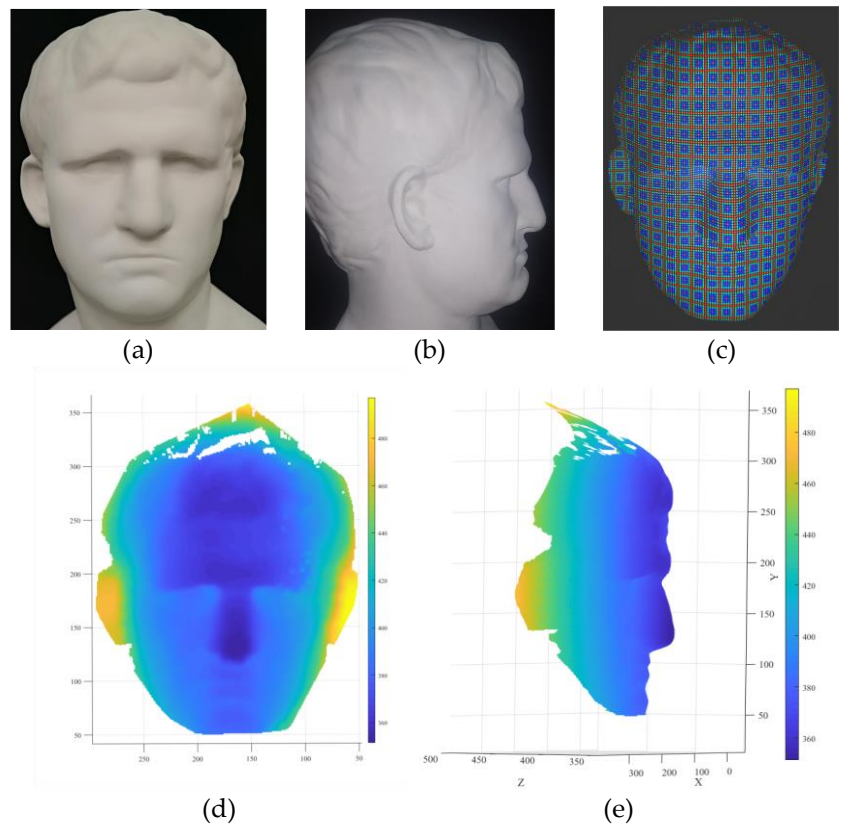

**Figure 20.** Results of reconstructing a face statue; **(a)** the face statue without structured light pattern in front face view; **(b)** the face statue without structured light pattern in side face view; **(c)** the face statue with structured light pattern in front face view; **(d)** the reconstructed face statue in front face view; **(e)** the reconstructed face statue in side face view.

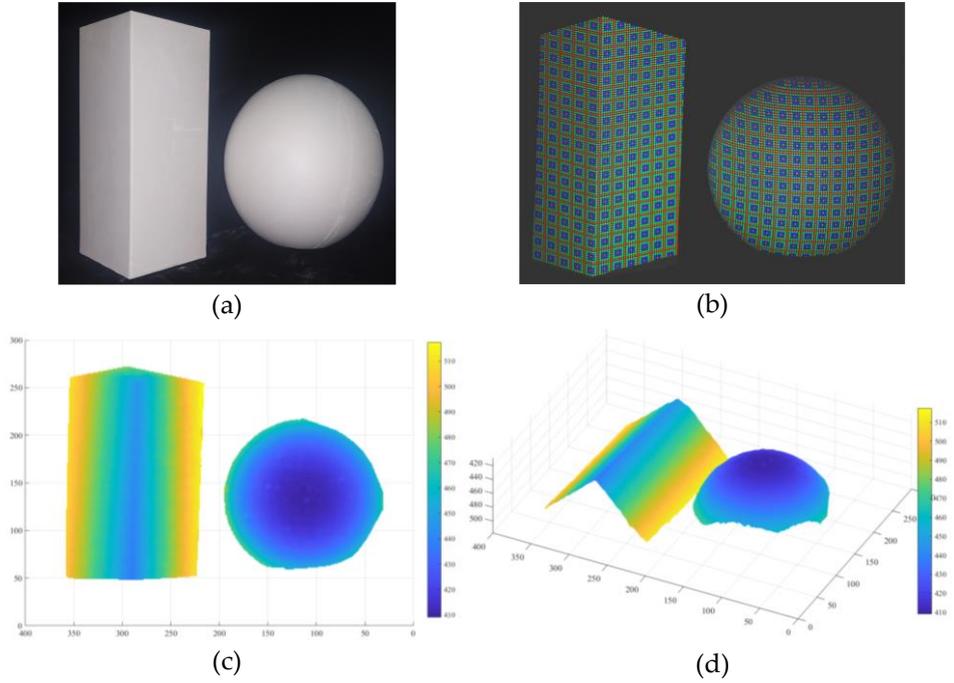

(a) (b) (c) (d)

**Figure 21.** Results of reconstructing two separate objects together; **(a)** the two objects without structured light pattern; **(b)** the two objects with structured light pattern; **(c)** the reconstructed two objects in view 1; **(d)** the reconstructed two objects in view 2.

We use several static objects to test the proposed approach further. The first static object is the cone with cylinder and the reconstruction results are shown in Figure 19. The second static object is a face statue and the reconstruction results are shown in Figure 20. We also tested the proposed approach to reconstruct two separate objects and the reconstruction results are shown in Figure 21. As can be seen, all the reconstructed objects appear highly accurate.

Besides the static objects, we also use the dynamic hand to test the proposed approach. The dynamic hand with different gestures and is reconstructed by the proposed approach continuously. Some of the reconstruction results are shown in Figure 22 and the whole reconstruction results of the dynamic hand is shown in the attached video (Visualization 1). It is seen from all the reconstruction results shown in this paper that the proposed approach could reconstruct both the static objects and the dynamic objects by single-shot robustly.

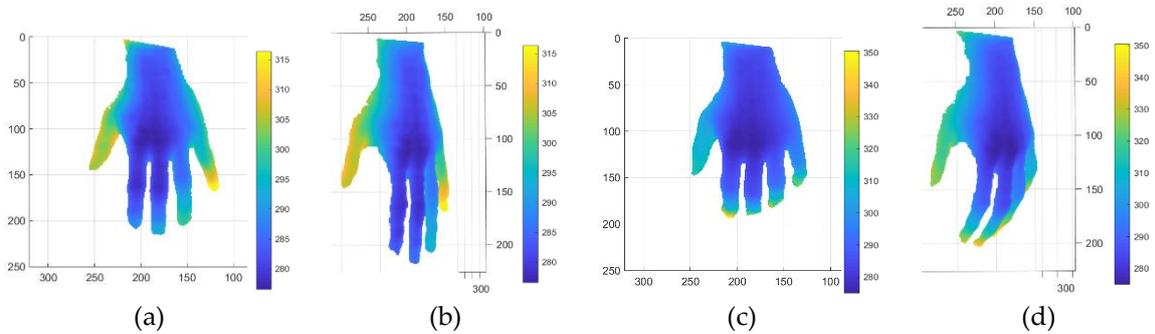

(a) (b) (c) (d)

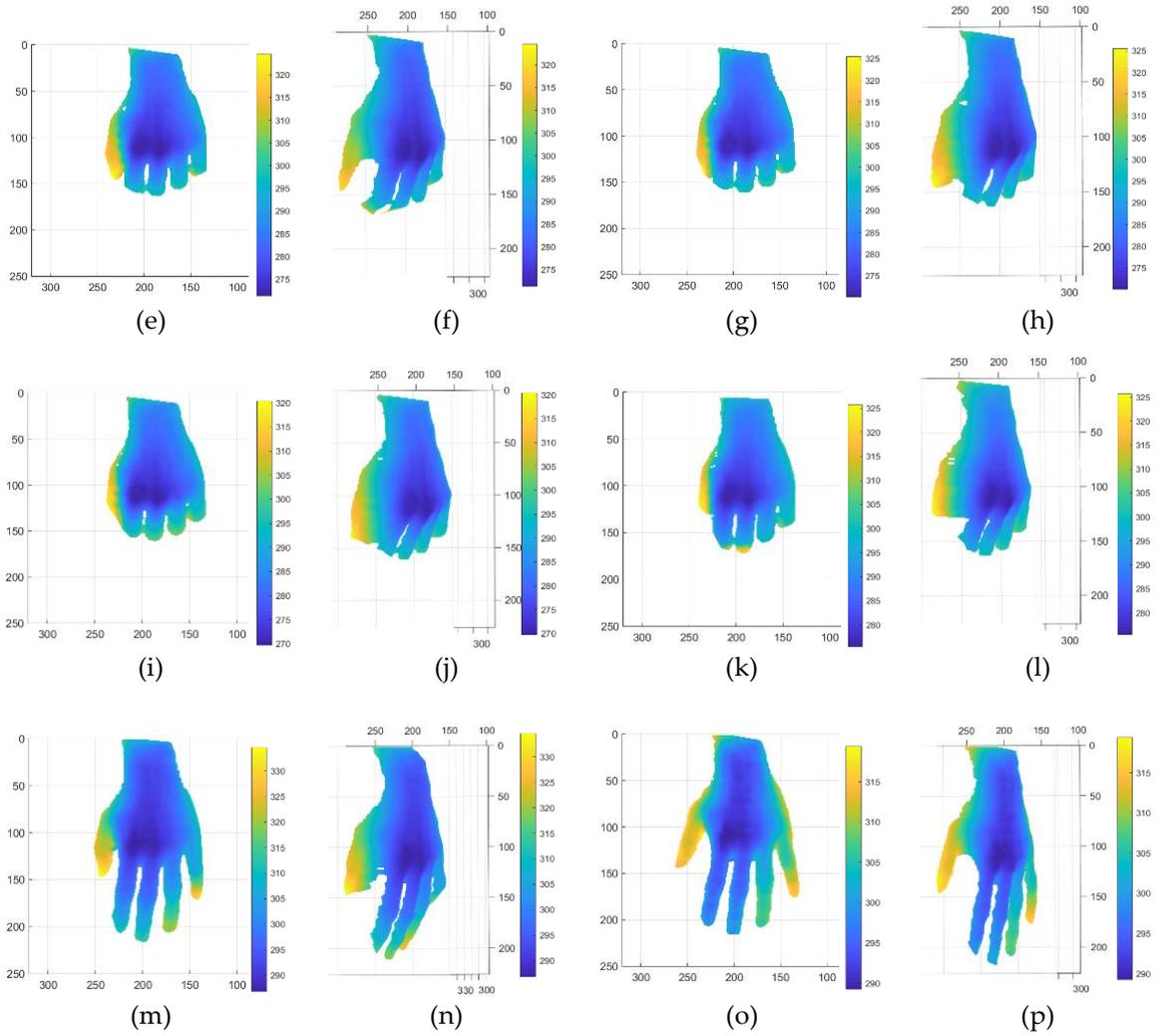

**Figure 22.** Results of reconstructing the dynamic hand by the proposed method; **(a)** the reconstructed hand with gesture 1 in view 1; **(b)** the reconstructed hand with gesture 1 in view 2; **(c)** the reconstructed hand with gesture 2 in view 1; **(d)** the reconstructed hand with gesture 2 in view 2; **(e)** the reconstructed hand with gesture 3 in view 1; **(f)** the reconstructed hand with gesture 3 in view 2; **(g)** the reconstructed hand with gesture 4 in view 1; **(h)** the reconstructed hand with gesture 4 in view 2; **(i)** the reconstructed hand with gesture 5 in view 1; **(j)** the reconstructed hand with gesture 5 in view 2. **(k)** the reconstructed hand with gesture 6 in view 1; **(l)** the reconstructed hand with gesture 6 in view 2. **(m)** the reconstructed hand with gesture 7 in view 1; **(n)** the reconstructed hand with gesture 7 in view 2. **(o)** the reconstructed hand with gesture 8 in view 1; **(p)** the reconstructed hand with gesture 8 in view 2.

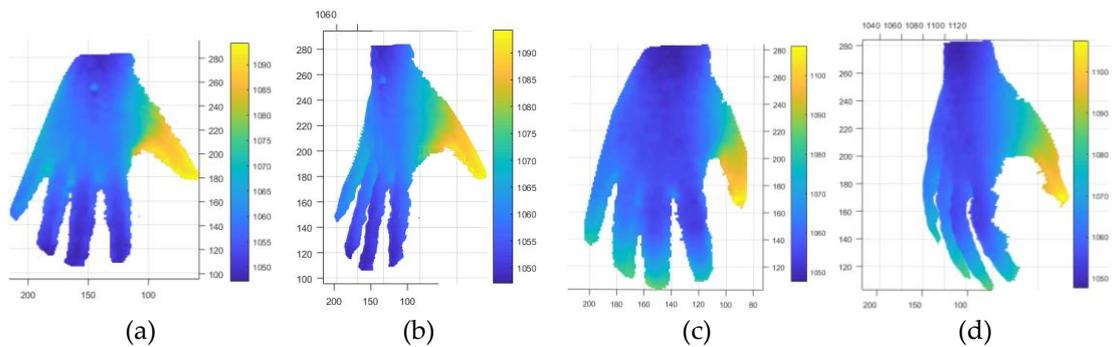

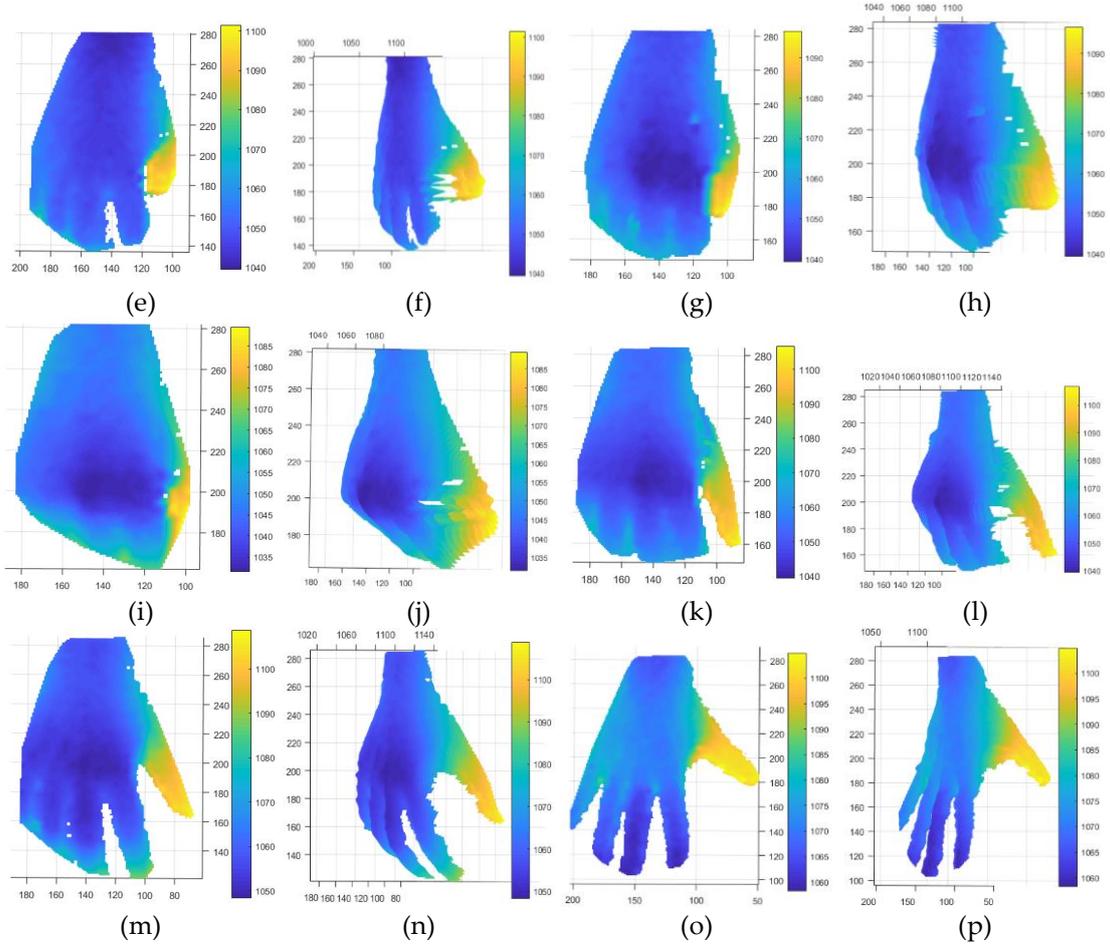

**Figure 23.** Results of reconstructing the dynamic hand by the method in [8]; **(a)** the reconstructed hand with gesture 1 in view 1; **(b)** the reconstructed hand with gesture 1 in view 2; **(c)** the reconstructed hand with gesture 2 in view 1; **(d)** the reconstructed hand with gesture 2 in view 2; **(e)** the reconstructed hand with gesture 3 in view 1; **(f)** the reconstructed hand with gesture 3 in view 2; **(g)** the reconstructed hand with gesture 4 in view 1; **(h)** the reconstructed hand with gesture 4 in view 2; **(i)** the reconstructed hand with gesture 5 in view 1; **(j)** the reconstructed hand with gesture 5 in view 2. **(k)** the reconstructed hand with gesture 6 in view 1; **(l)** the reconstructed hand with gesture 6 in view 2. **(m)** the reconstructed hand with gesture 7 in view 1; **(n)** the reconstructed hand with gesture 7 in view 2. **(o)** the reconstructed hand with gesture 8 in view 1; **(p)** the reconstructed hand with gesture 8 in view 2.

To verify the advantage of the block based dot pattern designed in this study over the line based dot pattern designed in [8], the same stereo vision system was used to reconstruct the dynamic hand with the method and the line based dot pattern designed in [8]. The qualitative results are shown in Fig. 23 for comparison with the results shown in Fig. 22. As can be seen, the results by the proposed approach look much better than the results by the method proposed in [8]. The advantages of the block based dot pattern proposed in the paper over the line based dot pattern proposed in [8] include: (1), when one point in the block is not matched correctly, it will be rectified by the iterative dot matching method proposed in this paper. On the contrary, when one point in the line is not matched correctly, the whole line will be incorrectly matched. As a result, the proposed pattern is more resistant to occlusion problems. (2), the dot extraction method in this paper is based on extreme value detection while the dot extraction method proposed in [8] is based on iterative threshold selection. The size of the designed dot in this paper is much smaller that of the designed dot in [8]. As a result, the dots used by the block pattern for reconstructing the same object are at least four times more than the dots used by the line pattern in [8].

To verify the accuracy advantage of the proposed approach further, we also compare the accuracy of the proposed approach with that of state of the art method [4] in Table 3. In [4], Kinect v1 whose working principle is light coding instead of the time of flight was used for accuracy comparison and the radius of the measured ball is 20 mm. In this paper, Kinect v2 whose working principle is time of flight is used for accuracy comparison and the radius of the measured ball is 84.68 mm. From the quantitative comparisons shown in Table 3, it is seen that the accuracy of the proposed approach is significantly more accurate than state of the art methods.

**Table 3.** Comparison of the accuracy of the proposed approach with that of the method in [4].

| Methods\Objects | error (mm) | error (percentage) |
|---|---|---|
| Kinect V1 in [4] | 0.845/20 mm | 4.22% |
| Laser speckle method in [4] | 0.392/20 mm | 1.96% |
| Kinect V2 in this paper | 0.92/84.68 mm | 1.09% |
| Proposed | 0.336/84.68 mm | 0.4% |

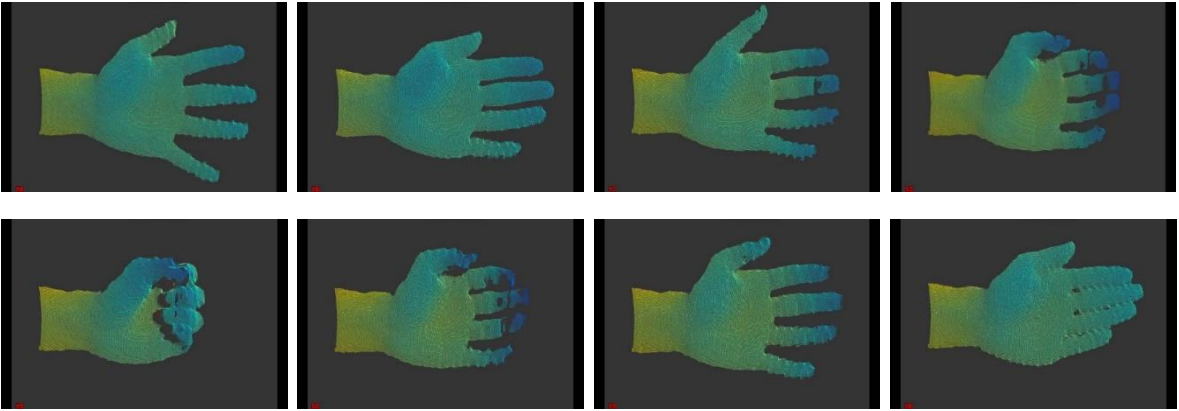

**Figure 24.** Reconstructed hand by the method in [7].

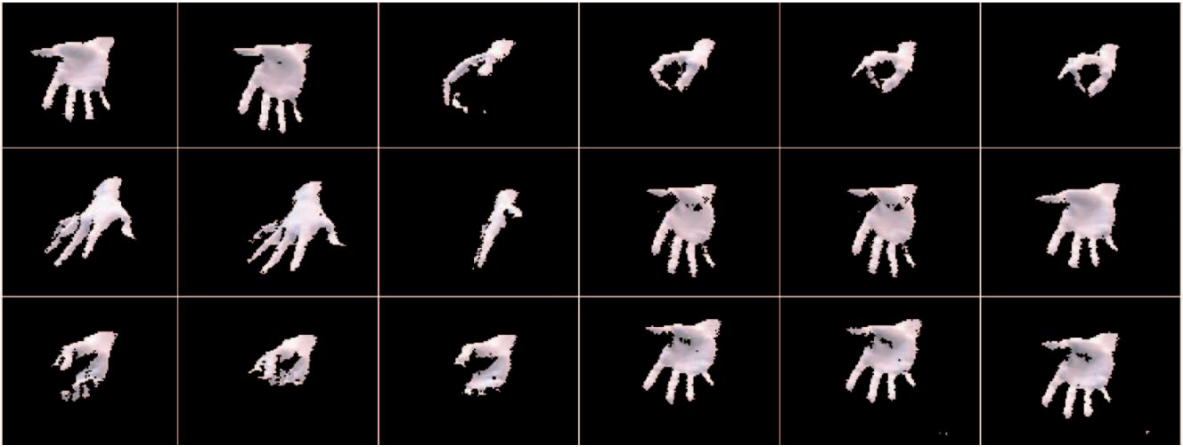

**Figure 25.** Reconstructed hand by the method in [26].

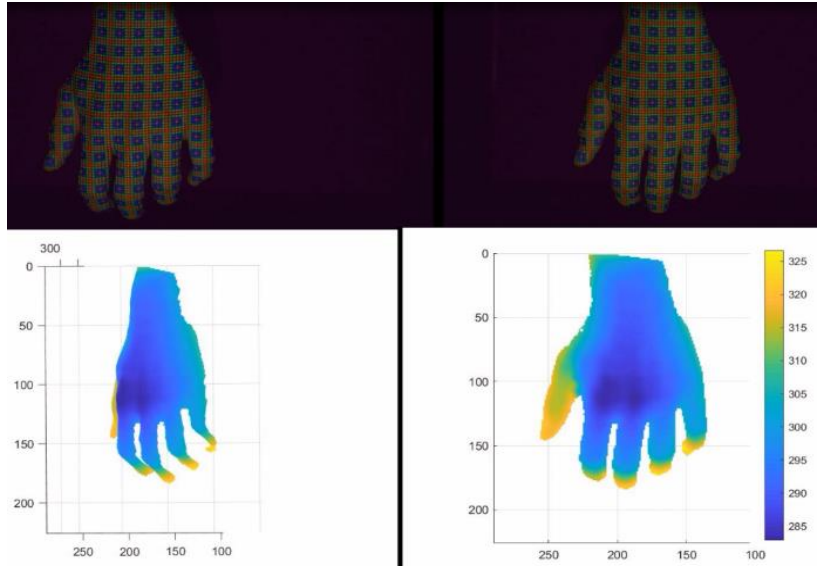

**Figure 26.** Demonstration of the reconstructed hand motion by the proposed approach with one selected frame of the attached video (See Visualization 1).

It has remained challenging for three-dimensional measurement techniques to reconstruct complex 3D motions, such as a dynamic hand. In [7], the authors tried to verify the effectiveness of their method by reconstructing a dynamic hand. Figure 24 shows eight frames of the attached video in [7]. As can be seen, the method in [7] generated many black holes. In [26], the authors also tried to evaluate their method by reconstructing a dynamic hand. Figure 25 shows eighteen frames of the reconstructed hand in [26]. As can be seen, the reconstructed hands are worse than those reconstructed in [7]. To compare the proposed approach with that proposed in [7] and [26], we also attach a video of reconstructing a dynamic hand as Visualization 1. Figure 26 shows one frame of the attached video. The top two images are the acquired pattern images by the left camera and the right camera respectively. The bottom two images show the reconstructed hand in two different view angles. As can be seen, the proposed approach is not affected by black holes when doing similar motions. Please note that this study focuses on reconstructing a complex and dynamic 3D shape instead of comparing the reconstruction accuracy with state of the art methods extensively because the reconstruction accuracy is determined by the ray intersection principle instead of the proposed pattern and the proposed matching method.

Similar to structured light technology, the proposed approach is easily affected the external light source. As a result, it achieves the best performance in the dark. In addition, when the reconstructed object has some specific textures, the performance of the proposed approach will also be affected. When the reconstructed surface is rugged instead of smooth, the proposed approach will perform poorly due to its interpolation operation. All these factors will limit the application of the proposed approach significantly and thus will be studied in our future research. Currently, the application field of the proposed approach is similar to or wider than that of the structured light method because its superiority in reconstructing 3D motions. The proposed approach could be used for all of the applications of structured light techniques described in [27]. However, none of the techniques described in [15, 27] could be used to reconstruct a dynamic hand robustly yet. As reviewed in [15, 28], the structured light methods that project a series of patterns for 3D reconstruction are affected by the motion-caused vibration noise when reconstructing the dynamic shape. On the contrary, active stereo vision that inherits the advantages of both structured light and stereo vision only needs to project one pattern for 3D reconstruction. As a result, active stereo vision is a much better choice for reconstructing the dynamic shape than most structured light techniques.

## 4. Conclusions

In this paper, a RGB dot pattern based active stereo vision 3D imaging approach is proposed to reconstruct both the static objects and the dynamic objects with single-shot. Specifically, a novel RGB dot pattern is designed, an effective dot extraction method is proposed and an effective active stereo vision matching method is proposed. Experimental results verified the effectiveness of all the proposed methods and especially verified the ability of the proposed 3D imaging approach in reconstructing 3D motions. The qualitative results showed that the reconstructed 3D hands by the proposed approach are significantly more complex and more complete than those reconstructed by state of the art methods. In addition, the quantitative accuracy comparison showed that the ray intersection based stereo vision is more accurate than disparity based stereo vision in measuring 3D shapes. The quantitative results showed that the proposed approach is significantly more accurate than state of the art real-time 3D measurement products: Microsoft Kinect V1 and Kinect V2. When the proposed approach and two of our previously proposed structured light methods were implemented with the same hardware, the proposed approach also achieved slightly better accuracy.

**Acknowledgments:** We thank the anonymous reviewers for their insightful comments.